%% file: main.tex
\documentclass[a4paper,UKenglish,cleveref,pdfa]{lipics-v2021}

\hideLIPIcs

\author{Ian Miguel}%
{School of Computer Science, University of St Andrews, UK}%
{ijm@st-andrews.ac.uk}%
{https://orcid.org/0000-0002-6930-2686}%
{EPSRC grant EP/V027182/1}

\author{András Z. Salamon}%
{School of Computer Science, University of St Andrews, UK}%
{Andras.Salamon@st-andrews.ac.uk}%
{https://orcid.org/0000-0002-1415-9712}%
{}

\author{Christopher Stone}%
{School of Computer Science, University of St Andrews, UK}%
{cls29@st-andrews.ac.uk}%
{https://orcid.org/0000-0002-9512-9987}%
{EPSRC grant EP/V027182/1}

\title{Automating Reformulation of Essence Specifications via Graph Rewriting
}

\authorrunning{I. Miguel and A.\,Z. Salamon and C. Stone}

\Copyright{Ian Miguel and András Z. Salamon and Christopher Stone}



\category{}

\relatedversion{}

\nolinenumbers

\usepackage{graphicx} 
\usepackage{xspace}
\usepackage{listings}
\lstdefinestyle{plain}{numbers=none,language=}

\usepackage{svg}
\usepackage{todonotes}
\usepackage{pdfpages}

\hypersetup{breaklinks=true,colorlinks,linkcolor=black,urlcolor=blue,citecolor=black}

\newcommand{\essence}{\textsc{Essence}\xspace}
\newcommand{\eprime}{\textsc{Essence Prime}\xspace}
\newcommand{\conjure}{\textsc{Conjure}\xspace}

\newcommand{\code}[1]{{\small\texttt{{#1}}}}

\begin{document}

\maketitle

\begin{abstract}

Formulating an effective constraint model of a parameterised problem class is crucial to the efficiency with which instances of the class can subsequently be solved. It is difficult to know beforehand which of a set of candidate models will perform best in practice. This paper presents a system that employs graph rewriting to reformulate an input model for improved performance automatically. By situating our work in the \essence abstract constraint specification language, we can use the structure in its high level variable types to trigger rewrites directly. We implement our system via rewrite rules expressed in the Graph Programs 2 language, applied to the abstract syntax tree of an input specification. We show how to automatically translate the solution of the reformulated problem into a solution of the original problem for verification and presentation. We demonstrate the efficacy of our system with a detailed case study.

\end{abstract}


\section{Introduction}

Formulating an effective constraint model of a problem of interest is crucial to the efficiency with which the problem can subsequently be solved~\cite{freuder2018:progress}. It is difficult to know beforehand which of a set of candidate models will perform best in practice. In this paper we present our work in progress on a system that reformulates an input model to improve performance automatically. It differs from some earlier work on automated model reformulation, such as {\sc CGRASS}~\cite{frisch2002cgrass}, Tailor~\cite{gent2007tailoring}, the work of Bessiere et al.~\cite{Bessiere2007:learning} to learn implied global constraints, or Savile Row~\cite{savilerow}, in that it reformulates a model of a parameterised problem class rather than individual problem instances. This has the advantage that the effort made to reformulate the model is amortised over all the instances of the class that are to be solved, rather than having to be paid back during the solution of a single instance. Furthermore, by implementing reformulation as forward-chaining sound rewrite rules, we avoid the need for external verification in existing class-based reformulation~\cite{colton2001:constraint, charnley2006:automatic, leo2022globalizing}.
Here we build on our previous proposal for a system~\cite{Miguel2023:towards}, demonstrating how to apply graph transformation techniques~\cite{icgt2024} to achieve automated rewriting of \essence specifications.

In contrast to recent work by Leo et al.~\cite{Leo2024:automatic}, who reformulate a low level model of a problem class, we situate our work in the \essence abstract constraint specification language~\cite{frisch2008:essence}. The advantage of this approach is that the structure apparent in a concise abstract specification can be used to trigger and guide reformulation. For example, \cref{lst:before} presents the k-fold colouring problem we will use in our case study. Here the single abstract decision variable is a binary relation. Having that information directly in the variable type, as opposed to reconstructing it from a constraint model-level representation likely composed of a constrained matrix of more primitive decision variables, is a significant aid to reformulation. Furthermore, a single reformulated specification can be refined into a variety of both models and solving paradigms, allowing us to gain a fuller picture of performance. We implement our system via rewrite rules expressed in the Graph Programs 2 language \cite{campbell2020improved,plump2017imperative}, applied to the abstract syntax tree of an input specification. Given the small size of \essence specifications and the efficiency of the GP2 system in applying graph rewriting rules, the rewriting process has negligible cost. When the type of the decision variable is transformed, the solution must be converted to the original type for verification and presentation to the user. A constraint solver can be employed to solve the solution re-assignment specification straightforwardly. We demonstrate this process even in the case of the nested types supported by \essence.

Our work makes the following primary contributions:
\begin{itemize}
\item Automated class-level model reformulation via a library of graph transformation rewrite rules.
\item Inverse rewriting of solutions for presentation and verification.
\item A case study using k-fold graph colouring.
\end{itemize}

\begin{lstlisting}[caption={\essence specification of $k$-fold graph colouring, a variant in which each node in a given graph must be assigned $k$ colours with no pair of nodes connected by an edge sharing a colour.},label={lst:before},frame=tlrb]{Name}
$ k-fold graph colouring with k=coloursPerNode, out of numberColours
given n : int
letting vertices be domain int(0..n-1)
given edges : relation (irreflexive) of ( vertices * vertices )
given numberColours : int(1..)
given coloursPerNode : int(1..)
letting colours be domain int(1..numberColours)
find colouring : relation (size n*coloursPerNode) of (vertices * colours)
such that

$ endpoints of edges do not share colours
forAll (u,v) in edges .
     (forAll colourAssignment in colouring .
        (colourAssignment[1] = u) -> !((v,colourAssignment[2]) in colouring)),

$ enforce number of colours per node
forAll u : vertices .
     coloursPerNode = (sum colourAssignment in colouring .
        toInt(colourAssignment[1] = u))
        
\end{lstlisting}

\section{Rewriting via graph transformation}

Reformulation can be conceptualised in various ways. In this study, we have chosen to pursue a specific framework for reformulation based on rewriting. The specification at each stage of reformulation can be represented as an abstract syntax tree (AST), from which the specification can be recovered without loss of information. ASTs are commonly modelled as trees, but it is also possible to replace common subtrees in the AST with pointers, transforming the data structure into the form of a directed acyclic graph. In either case, ASTs can be represented as graphs consisting of a set of labelled vertices and a set of directed and labelled arcs between vertices. A node in the AST corresponds to a labelled vertex. A graph rewriting system nondeterministically matches a pattern graph to the target graph; if a match is found, the matched part of the target graph is replaced by a different graph according to the rewrite rule. The details of matching and the types of rewriting may vary, but the graph rewriting paradigm is general enough to be Turing-complete~\cite{habel2001computational} and can thus capture the desired reformulation sequences. Each reformulation in a sequence of reformulations can be considered as a step taken by a graph rewriting system acting on the AST as the target graph being rewritten. Each reformulation is expressed by a graph rewrite rule or collection of rules. In this study, we have employed the graph rewriting language GP2~\cite{plump2017imperative} to perform rewriting on the abstract syntax tree representing a specification. The GP2 system provides a flexible language for expressing graph rewriting rules and includes an efficient implementation of the rewriting engine optimised for sparse large graphs~\cite{campbell2020improved}. The GP2 language can be used to write graph transformation rules, which are compiled using the GP2 engine and become stand-alone C programs that can be executed independently. In the GP2 language, graphs are defined using a set of nodes that are tuples of indices and labels followed by a set of edges that are quadruples of indices, source nodes, target nodes and labels.

\section{Reformulating Essence specifications for performance}

\begin{figure}[htbp]
\centering
\vspace{-10pt}
\def\svgwidth{\columnwidth}
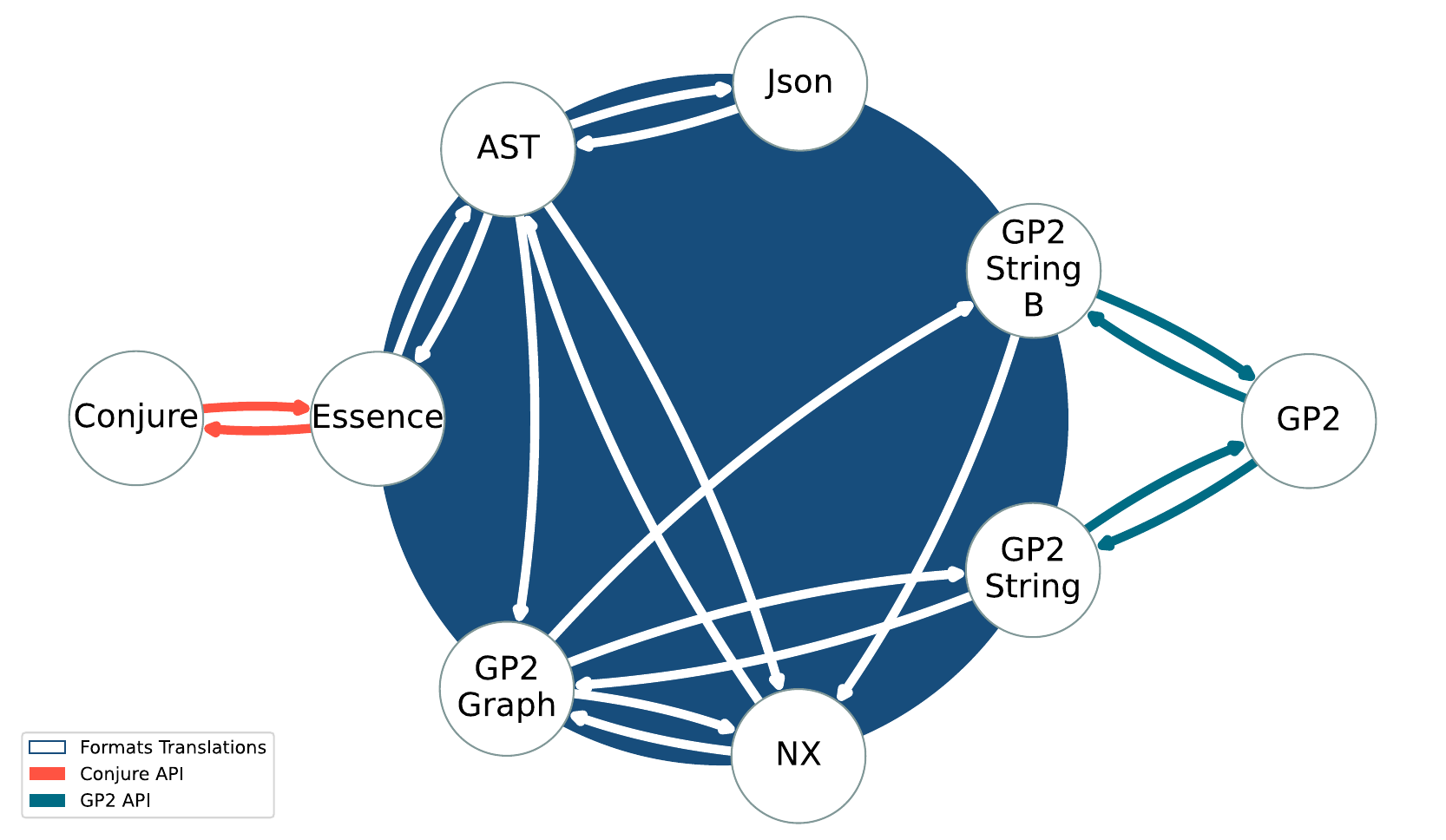
  \caption{Graph showing format conversions and APIs. The white arrows within the blue area are parts of our system of conversion and the green arrows show interactions with GP2.}
  \label{fig:formats_graph}
\end{figure}

The \essence language allows problems to be expressed concisely using high level types such as sets, multisets, functions, relations, and partitions, which may be nested -- sets of functions, multisets of relations, etc. \essence specifications are typically solved by a process of {\em refinement} via the {\sc Conjure} tool~\cite{Akgun2022:Conjure}, where the high level types are transformed into collections of more primitive variables, such as matrices of integers, in the solver-independent \eprime language~\cite{nightingale2016essence}. The {\sc Savile Row} tool~\cite{savilerow} further transforms a model in \eprime into input suitable for a particular constraint, SAT, or SMT solver.

This complex chain of refinement and transformation is sensitive to the \essence specification with which it starts, and so an  initial specification may not always lead to optimal solving performance. Therefore, reformulating the specification becomes necessary to improve the efficiency of the solving process. This is often where modellers, through their expertise, can obtain the largest gains in efficiency~\cite{Regin2013:ACP}. One of our objectives is to help automate the application of reformulation rules, sourcing them both from expert modellers or those found via other search methods and building a library of rewriting rules. An advantage of reformulating \essence is that even a small change can produce a marked change once refinement and parameter substitution has taken place.

The gains provided by reformulations are not only limited to faster solving time but, crucially, can lead to smaller memory footprints across the subsequent refinement steps. In some cases this allows problems to be solved that in their original form may not be able to reach the solver or that could crash during solving.

In practice, a specification written in \essence is turned into GP2 form by a series of conversions and translations by following the shortest path in a conversion graph pictured in Figure~\ref{fig:formats_graph}. We extend our previous work~\cite{icgt2024},
which contains formal definitions for each intermediate form, by adding a different representation of the AST once in GP2 form. In our approach, instead of compounding symbols and their grammatical information into the labels of the GP2 nodes, we add an extra branch to each node separating the two. Once in GP2 form, any of the available precompiled rules can be applied to the AST of the \essence specification. In this particular study, we are interested in a rewrite rule designed to showcase the potential of high-level reformulations making use of relaxation, auxiliary domains, type nesting and constraint rewrites.
A fragment of our rule set can be seen in \cref{lst:ruleGP2}, where the left-hand side of the rule is designed to match the portion of the AST in which a relation is declared. The rule takes three parameters, \code{specName}, \code{decisionVariableName} and \code{findPos}, two strings and an integer, respectively, capturing the fact that the labels on those nodes can be any value and can be reused during the rewriting step. Labels in quotes such as \code{"find"} and \code{"relation"} must be matched exactly. Nodes in the interface are nodes that will be preserved. In this case, all the nodes are preserved, and only one label is changed by attaching the flag \code{\# red} that will trigger subsequent rules.

\begin{lstlisting}[caption={Tagging a decision variable node implemented in GP2.},label={lst:ruleGP2}]
Main = tagRelationDecisionVariable

tagRelationDecisionVariable(specName,decisionVariableName:string;findPos,n:int)
\\ left-hand side of the rule
[ 
    (n0, specName) (n1, "find") (n2, decisionVariableName) (n3, "relation")
|
    (e1, n0, n1, findPos) (e2, n1, n2, 1) (e3, n2, n3, 1)
]
=>
\\ right-hand side of the rule
[
    (n0, specName) (n1, "find") (n2, decisionVariableName) (n3, "relation"# red)
|
    (e1, n0, n1, findPos) (e2, n1, n2, 1) (e3, n2, n3, 1)
]
interface = {  n0,n1, n2, n3 }
\end{lstlisting}

\section{Case Study: \texorpdfstring{$k$-fold}{k-fold} fractional graph colouring}

In order to demonstrate our reformulation system in operation we present a case study with {\em fractional graph colouring}, which is a generalisation of the graph colouring problem where multiple distinct colours are assigned to each vertex of a graph such that pairs of neighbouring vertices have no colours in common. In the literature it can be found under various different names: Graph Multicoloring Problem, Generalised Graph Colouring Problem, k-Fold Graph Colouring, Fractional Graph Colouring, or Fractional Vertex Colouring. It has been used in the past to showcase constraint programming techniques~\cite{gualandi2012exact,prestwich2008generalised} and it is commonly found in the context of scheduling applications~\cite{jansen2004preemptive}. We will refer to the integer value $k$ in $k$-fold as the number of colours each node must be assigned, and we will refer to the number of colours available explicitly.

In our case study we take as the input specification that is shown in \cref{lst:before}, which employs a binary relation to encode the graph and a binary relation for the decision variable linking each node with the colours assigned to it. This is a natural choice as graphs are commonly modelled using relations, and more than one colour has to be mapped to each node. Two constraints enforce that adjacent vertices do not share the same colour and that the correct number of colours is assigned to each vertex (via a common pseudo-Boolean counting pattern).

From this initial specification, it is possible to activate a sequence of rewriting rules that reformulate the specification with the objective of improving its performance. It can be noted that the number of colours necessary for each vertex is the same for all vertices, the colours assigned to each vertex could be expressed as a set of fixed size, eliminating the need for the counting constraint. This means that one rewrite of the rule should be able to introduce auxiliary domains and another eliminate a constraint.
It follows that the decision variable be could turned into a total function that maps each vertex to a set of colours, further removing the need for a quantification, as referring to a set allows referring to all the colours of a specific node in one simpler statement. This requires a rewrite that turns the relation into a function and modifies the existing constraint.

The graph transformation rules proceed as follows:
\begin{enumerate}
    \item A rewrite rule matches a decision variable of arbitrary name and relation type, that is also mentioned in a quantification with a counting pattern, plus a second relation where the binary relation is over the same object (which is common for graphs) and a quantification over it.
    Occurrences of the decision variable \code{colouring} are bold blue in \cref{fig:coloured_graph1}.
    \item The rule tags the decision type's node, the head of the subtree of the constraint to be eliminated and the inner quantification of the colour that checks that adjacent vertices do not have the same colours.
    The tagged nodes are highlighted in red in \cref{fig:coloured_graph1}.
    \item Then, an additional rule propagates the tags in each subtree. 
    \item Then, a rule is applied until possible to match each pair of tagged nodes; the deeper of the two is removed, and so is the edge between them. This is to avoid deleting a node while leaving a dangling edge.
    \item The remaining tagged nodes are used to glue in the desired subtrees, while removing the single node's tags. Three rules match the specific tagged nodes and add the auxiliary domain, the new decision variable and the reformulated constraint.
    \item The remaining tagged nodes are removed, together with their parent edge.
\end{enumerate}

\begin{figure}[ht]
    \centering
    \includegraphics[width=\columnwidth]{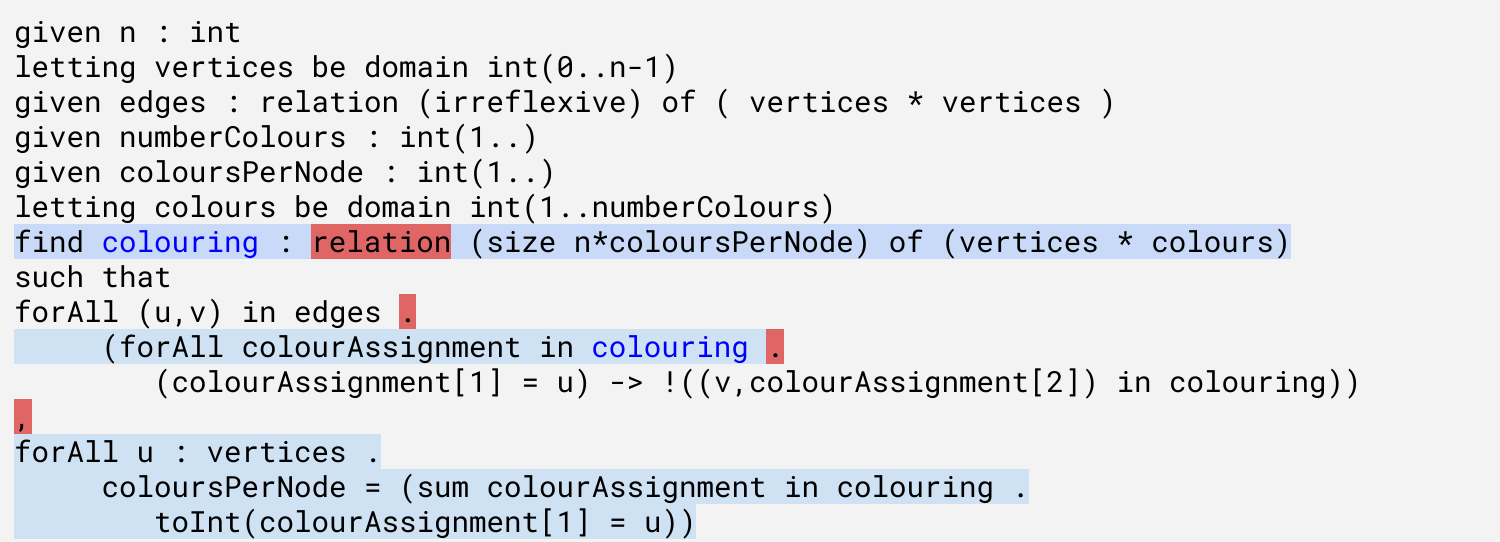}
    \caption{Parts of the specification that are interacted with during rewrite. Matched subtrees are highlighted in light blue. Names that can be used as free parameters are coloured in dark blue. Tagged nodes are highlighted in red.}
\end{figure}

\begin{lstlisting}[caption=Rewritten specification.,label={lst:after},frame=tlrb]{Name}
given n : int(0..100)
letting vertices be domain int(0..n-1)
given edges : relation (irreflexive) of ( vertices * vertices )
given numberColours : int(1..n)
given coloursPerNode : int(1..n)
letting colours be domain int(1..numberColours)

letting coloursSet be domain set (size coloursPerNode) of colours

find colouring : function (total) vertices --> coloursSet
such that
forAll (u,v) in edges .
     colouring(v) intersect colouring(u) = {}
\end{lstlisting}

This procedure adds an auxiliary domain of type \code{set} that will host the colours, a new decision variable of type \code{function} mapping the left-hand side of the relation to the auxiliary set, and shrinks the constraints to simply require an empty intersection of two sets. The final product can be seen in \cref{lst:after}.

\subsection{Empirical evaluation of the reformulation}

In order to test the performance of the reformulated abstract specification we have developed an instance generator that produces input parameters; this uses a simple grid search spanning a range of values for $n$ nodes and $e$ edges for the graph to be coloured, $c$ colours available and $cpn$ colours per nodes enforced.
Values. $n$: 10-40, step size 10. $e$: $n^2 *20\%-80\%$, step size 5\%. $cpn$: 2-5, step size 1. $c$: $cpn$*4-6, step size 1.

\begin{figure}[ht]
\centering
\def\svgwidth{\columnwidth}
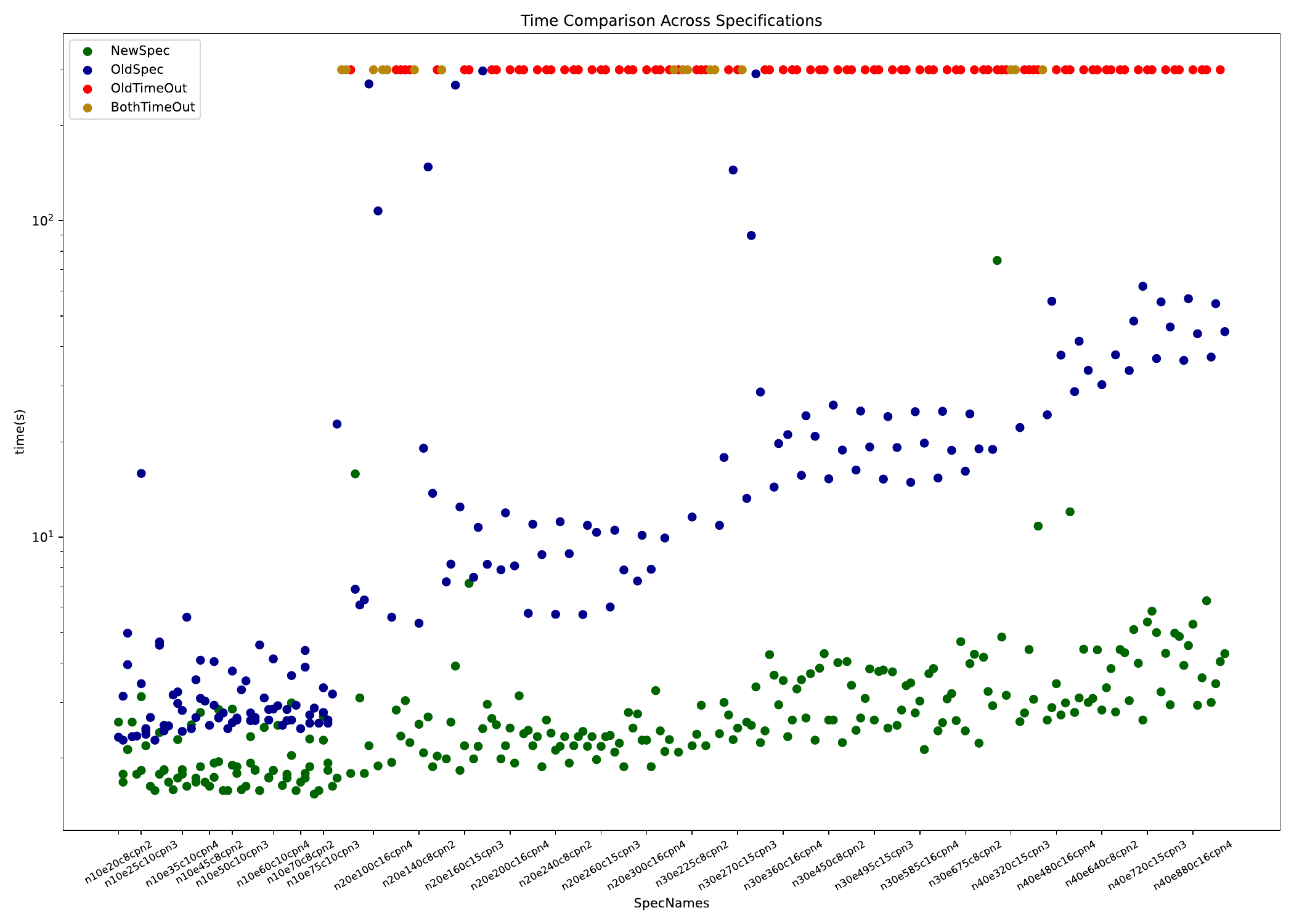
  \caption{Performance of the two specifications: we plot instances (with selected names listed along the horizontal axis), the reformulated specifications are green, the old specifications are blue, gold indicates both specifications time out, red indicates that the old specification times out.}
  \label{fig:specs_scatter}
\end{figure}

Each instance is solved using both models and the results are plotted in~\cref{fig:specs_scatter}.

\begin{figure}[ht]
\centering
\vspace{-10pt}
\def\svgwidth{0.75\columnwidth}
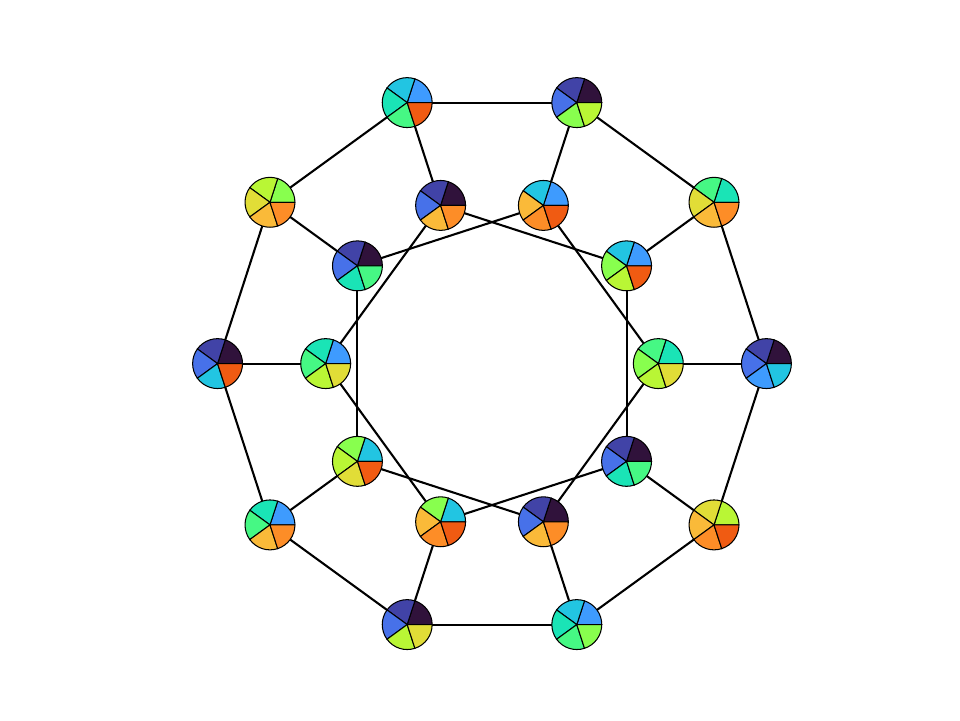
  \caption{5-fold colouring of the dodecahedral graph.}
  \label{fig:coloured_graph1}
\end{figure}

Our system can also translate graphs from the Python library NetworkX into \essence relations, which allows the conversion of any of its library of graphs into instances by supplying the $c$ and $cpn$ parameters. A small selection of iconic graphs has been used for additional testing, such as obtaining the 5-fold colouring of the dodecahedral graph in~\cref{fig:coloured_graph1}.

\section{Converting solutions to their original type}

When a reformulation changes the type of a decision variable, the solution produced by the solver will be of the transformed type. To verify that the solution of the reformulated specification is, in fact, a solution to the original problem, we must convert it back to its original type. To achieve this, we create a converter based on an \essence specification, which requires the solution of the reformulated problem as a parameter. The converter (illustrated in \cref{lst:converter}) includes all the variables declared in the original specification, along with all of its parameters. Additionally, it must have a decision variable with exactly the same name as the original specification, and a constraint mapping assignments of the new type to the old type. Solving this parameterised specification produces a solution that follows the structure of the original type that can be checked using \conjure with the \code{--validate-solution} option.

We timed the conversion of all the solutions obtained by solving the instances produced in the previous section. The worst-performing conversion took 0.56 seconds, with an overall average of 0.459 seconds. This overhead is significant only for the smallest instances.
\begin{figure}[ht]
    \centering
\includegraphics[width=\columnwidth]{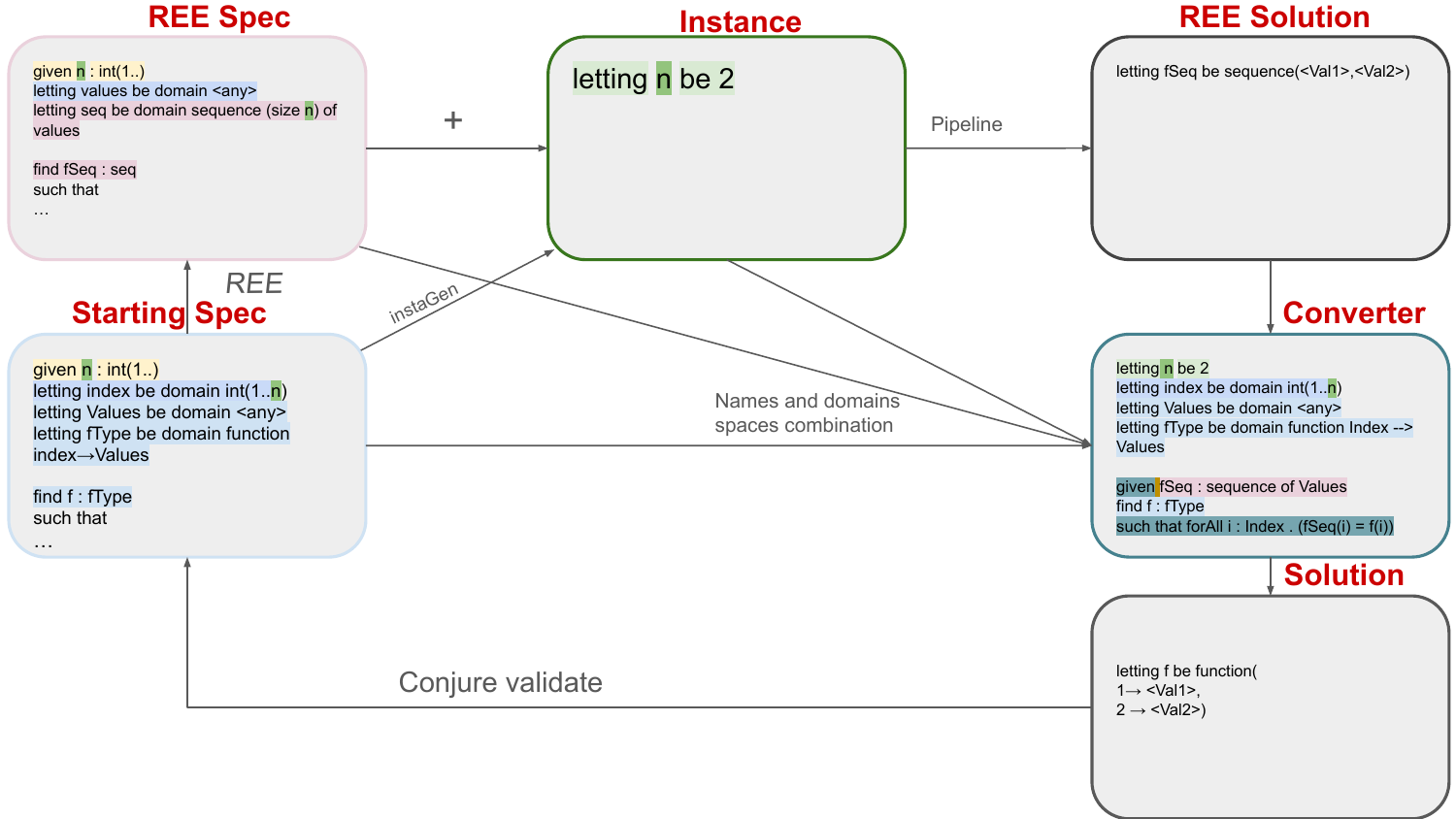}
    \caption{Specification of the solution's converter and production flow. In this simpler example a function indexed by integers is turned into a sequence and back again.}
\end{figure}

\begin{lstlisting}[caption=Solution Converter,label={lst:converter}]{Name}
given n : int(0..100)
given numberColours, coloursPerNode : int(1..n)
letting vertices be domain int(0..n-1)
letting colours be domain int(1..numberColours)
letting coloursSet be domain set (size coloursPerNode) of colours
given solution : function (total) vertices --> coloursSet
find colouring : relation (size n*coloursPerNode) of ( vertices * colours )
such that forAll item in defined(solution) . forAll colour in solution(item) . colouring(item, colour)
\end{lstlisting}


\section{Conclusion}

In this paper we have demonstrated an automated system for the reformulation of \essence specifications. Our system, implemented in the Graph Programs 2 language, reformulates specifications of parameterised problem classes, giving the key advantage of reformulation effort being amortised over the set of instances from the class to be solved. 
High-level \essence specifications are concise and highly structured, aiding rewriting. Furthermore, they are not tied to any particular solving paradigm, hence specification reformulations may be tested against a variety of modelling and solving choices.

{\bf Future work.}
Thanks to this foundational work, we have created the basis necessary to develop a system where the rewrite rules are produced automatically, or semi-automatically, starting from pairs of specifications that express possible reformulations. Automating the creation of rewrite rules would not only free the developer from the need to write GP2 programs but also collect rewrites from past specifications that have been refined over time in order to populate a rewriting library by distilling previous expert knowledge sitting idle in existing catalogues and literature. Another enticing future direction is the production of rewrite rules using search methods, which could be applied to any arbitrary specification. The aim would be to produce reformulations that improve the current model under consideration and are reusable.

\bibliography{main.bib}
\end{document}

%% file: CP_2024map_svg-tex.pdf_tex
\begingroup%
  \makeatletter%
  \providecommand\color[2][]{%
    \errmessage{(Inkscape) Color is used for the text in Inkscape, but the package 'color.sty' is not loaded}%
    \renewcommand\color[2][]{}%
  }%
  \providecommand\transparent[1]{%
    \errmessage{(Inkscape) Transparency is used (non-zero) for the text in Inkscape, but the package 'transparent.sty' is not loaded}%
    \renewcommand\transparent[1]{}%
  }%
  \providecommand\rotatebox[2]{#2}%
  \newcommand*\fsize{\dimexpr\f@size pt\relax}%
  \newcommand*\lineheight[1]{\fontsize{\fsize}{#1\fsize}\selectfont}%
  \ifx\svgwidth\undefined%
    \setlength{\unitlength}{799.20001221bp}%
    \ifx\svgscale\undefined%
      \relax%
    \else%
      \setlength{\unitlength}{\unitlength * \real{\svgscale}}%
    \fi%
  \else%
    \setlength{\unitlength}{\svgwidth}%
  \fi%
  \global\let\svgwidth\undefined%
  \global\let\svgscale\undefined%
  \makeatother%
  \begin{picture}(1,0.58117191)%
    \lineheight{1}%
    \setlength\tabcolsep{0pt}%
    \put(0,0){\includegraphics[width=\unitlength,page=1]{CP_2024map_svg-tex.pdf}}%
  \end{picture}%
\endgroup%

%% file: dodeca_turbo_palette_svg-tex.pdf_tex
\begingroup%
  \makeatletter%
  \providecommand\color[2][]{%
    \errmessage{(Inkscape) Color is used for the text in Inkscape, but the package 'color.sty' is not loaded}%
    \renewcommand\color[2][]{}%
  }%
  \providecommand\transparent[1]{%
    \errmessage{(Inkscape) Transparency is used (non-zero) for the text in Inkscape, but the package 'transparent.sty' is not loaded}%
    \renewcommand\transparent[1]{}%
  }%
  \providecommand\rotatebox[2]{#2}%
  \newcommand*\fsize{\dimexpr\f@size pt\relax}%
  \newcommand*\lineheight[1]{\fontsize{\fsize}{#1\fsize}\selectfont}%
  \ifx\svgwidth\undefined%
    \setlength{\unitlength}{460.79998779bp}%
    \ifx\svgscale\undefined%
      \relax%
    \else%
      \setlength{\unitlength}{\unitlength * \real{\svgscale}}%
    \fi%
  \else%
    \setlength{\unitlength}{\svgwidth}%
  \fi%
  \global\let\svgwidth\undefined%
  \global\let\svgscale\undefined%
  \makeatother%
  \begin{picture}(1,0.75000003)%
    \lineheight{1}%
    \setlength\tabcolsep{0pt}%
    \put(0,0){\includegraphics[width=\unitlength,page=1]{dodeca_turbo_palette_svg-tex.pdf}}%
  \end{picture}%
\endgroup%

%% file: performance_scatter_plot_svg-tex.pdf_tex
\begingroup%
  \makeatletter%
  \providecommand\color[2][]{%
    \errmessage{(Inkscape) Color is used for the text in Inkscape, but the package 'color.sty' is not loaded}%
    \renewcommand\color[2][]{}%
  }%
  \providecommand\transparent[1]{%
    \errmessage{(Inkscape) Transparency is used (non-zero) for the text in Inkscape, but the package 'transparent.sty' is not loaded}%
    \renewcommand\transparent[1]{}%
  }%
  \providecommand\rotatebox[2]{#2}%
  \newcommand*\fsize{\dimexpr\f@size pt\relax}%
  \newcommand*\lineheight[1]{\fontsize{\fsize}{#1\fsize}\selectfont}%
  \ifx\svgwidth\undefined%
    \setlength{\unitlength}{1008bp}%
    \ifx\svgscale\undefined%
      \relax%
    \else%
      \setlength{\unitlength}{\unitlength * \real{\svgscale}}%
    \fi%
  \else%
    \setlength{\unitlength}{\svgwidth}%
  \fi%
  \global\let\svgwidth\undefined%
  \global\let\svgscale\undefined%
  \makeatother%
  \begin{picture}(1,0.71428571)%
    \lineheight{1}%
    \setlength\tabcolsep{0pt}%
    \put(0,0){\includegraphics[width=\unitlength,page=1]{performance_scatter_plot_svg-tex.pdf}}%
  \end{picture}%
\endgroup%